\documentclass[11pt]{article}
\usepackage{geometry}
\usepackage{coling2020}
\usepackage{graphicx}
\usepackage{times}
\usepackage{url}
\usepackage{latexsym}
\usepackage{microtype}
\usepackage{multirow}
\usepackage{multicol}
\usepackage{booktabs}
\usepackage{caption}
\usepackage{comment}

\colingfinalcopy 

\title{CUHK at SemEval-2020 Task 4: CommonSense Explanation, Reasoning and Prediction with Multi-task Learning}

\author{
  Hongru Wang$^{1}$, Xiangru Tang$^3$, Sunny Lai$^{1}$, Kwong Sak Leung*$^{1}$, \\
  \textbf{Jia Zhu}$^4$, \textbf{Gabriel Pui Cheong Fung}$^{2}$, \textbf{Kam-Fai Wong}$^{2}$ \\
  $^1$Department of Computer Science and Engineering, The Chinese University of Hong Kong \\
  $^2$Department of Systems Engineering and Engineering Management, The Chinese University of Hong Kong \\ 
  $^3$Institute of Computing Technology, Chinese Academy of Sciences \\
  $^4$School of Computer Science, South China Normal University \\
  {\tt \{hrwang, slai, ksleung\}@cse.cuhk.edu.hk } \\
  {\tt jzhu@m.scnu.edu.cn \{pcfung, kfwong\}@se.cuhk.edu.hk}
  }
\date{}

\begin{document}
\maketitle

\begin{abstract}
 This paper describes our system submitted to task 4 of SemEval
 2020: Commonsense Validation and Explanation (ComVE) which consists of three 
 sub-tasks. The challenge is to directly 
 validate whether the system can recognize natural language statements that make sense from those that do not, and also require to generate reasonable explanation. Based on BERT architecture with multi-task setting, we propose an  effective and 
 interpretable ``Explain, Reason and Predict" (ERP) system to solve the 
 three sub-tasks about commonsense: (a) Validation, and (c) 
 Explanation, (b) Reasoning, following the order of the competition. Inspired by cognitive studies of common sense, our system first 
 generate a reason or understanding of the sentences and then choose which one 
 statement makes sense, which is achieved by multi-task learning. The rational experiment validates our assumption and boost the performance. During the 
 post-evaluation, our system has reached 92.9\% accuracy in subtask A (rank 11),
 89.7\% accuracy in subtask B (rank 8), and BLEU score of 12.9 in subtask C (rank
 9)\footnote{All results before 29 April, 2020}.
\end{abstract}

\section{Introduction}
\label{chap:introduction}




How to integrate common sense into natural language models is attracting more and more attention. Common sense, as ordinarily conceived, present themselves as the aspect of the grammar of expressions and sentences on which their semantic properties and relations depend~\cite{asher1995toward}. And a critical difference of text understanding between humans and machines lies in the fact that humans can access commonsense knowledge while processing text, which helps them draw inferences about effects that are not mentioned in a paragraph. Therefore, it's a fundamental question on how to validate whether a system has a commonsense capability, and more importantly, let the system explain how it reasons using hidden facts.
Existing benchmarks measure commonsense knowledge indirectly and without explanation. Also, existing datasets test common sense indirectly through tasks that require extra knowledge, such as co-reference resolution, or reading
comprehension. They verify whether a system is equipped with common sense by testing whether it can give a correct answer to make the complete sentence reasonable. However, there are some restrictions on such benchmarks. First, they do not provide a straight quantitatively standard to measure sense masking
capability. Second, they do not explicitly identify the key factors required in a sense-making process. And also, they do not need the model to explain why it makes that prediction.

\blfootnote{*Corresponding author.}
\blfootnote{This work is licensed under a Creative Commons Attribution 4.0 International License. License details: http://creativecommons.org/licenses/by/4.0/}

Common sense reasoning require the agent or the model to utilize a world knowledge to take inferences or deep semantic understanding, not only the pattern recognition.

Some empirical analysis has been done previously for common sense reasoning, mainly focus on the form of question answering (QA) \cite{Talmor2019CommonsenseQAAQ}. 
But question-answering is hard to directly evaluate the commonsense in contextualized representations.
And there has been few work investigating commonsense in pre-trained language models~\cite{zhou2019evaluating}, such as ELMo~\cite{Peters:2018} and 
BERT~\cite{Devlin2019BERTPO}. Introduced by \cite{wang-etal-2019-make}, sense-making is a task to  tests whether a model can differentiate sense-making and non-sensemaking statements. Specifically, the statements typically differ only in one keyword which covers nouns, verbs, adjectives, and adverbs. There are two existing approaches that can address this problem, one simple way is to use  more commonsense knowledge can be learned from larger training sets~\cite{Wang2019ImprovingNL}.
On the other hands, some works~\cite{Lin2019KagNetKG}  focus on effectively utilizing external, structured commonsense knowledge graphs, such as  ConceptNet~\cite{Speer2016ConceptNet5A} and COMET ~\cite{Bosselut2019COMETCT}. Insipred by previous works, more researchers are trying to fuse commonsense knowledge and language model~\cite{forbes2017verb}, and apply them to downstream tasks~\cite{Zhong2019KnowledgeEnrichedTF}. Recently, a new hybrid approach has been proposed for common sense reasoning~\cite{He2019AHN}. The core idea behind it is multi-task learning~\cite{Liu2019MultiTaskDN}, which has been widely applied in natural language tasks~\cite{Liu2019ImprovingMD}.


But existing work in this area has been frustratingly slow, and much of the work is completely theoretical. The field might well benefit if commonsense argumentation were systematically described and evaluated. To tackle it, this system focuses on a benchmark to directly test whether a 
system can differentiate sentences that make sense from those
that do not make sense. Our results indicate that pre-trained models are not able to demonstrate well on the benchmark, and  some remaining 
cases demonstrating that human level is not achieved yet. Thus, we design a new
procedure to handle the commonsense challenge inspired by human cognition. It firstly explain its understanding of the given sentences by a language model, and induce the hidden common sense fact. And then, the 
explanation is used as a supplementary input to the prediction module. Still, we believe that our approach also can be applied to more challenging data sets.

The organization of this paper is as follows: in Section 2, we introduce the basic information about pre-trained language model and task definition. We then describe the framework of our model in Section 3. Empirical results are given and discussed in Section 4. And then we provide more exhaustive analysis for some bad cases that appeared at our experiment in Section 5. Finally, we conclude this survey and in Section 6.

\section{Preliminaries}

\subsection{Task Definition}
Formally, the dataset is composed of 10 sentences: two sentences ${s_1, s_2}$, three options ${o_1, o_2, o_3}$, three references ${r_1, r_2, r_3}$. The first two natural language statements that are two similar sentences but only one of them makes sense are used in subtask A called Validation, For the against-common-sense statement $s_1$ or $s_2$, we have three optional sentences $o_1$, $o_2$ and $o_3$ to explain why the statement contradicts common sense. The subtask B, named Explanation (Multi-Choice), requires that the model can identify the 
only correct reason from distractors. Finally, subtask C naming 
Explanation (Generation), asks the model to generate the reason why it 
does not make sense. Each statement is paired with three possible explanations $r_1$, $r_2$ and $r_3$ from different perspective\cite{wang-etal-2020-semeval}.

\textbf{Subtask A:} Unlike other classification problem, subtask A gives us two
statements $s_1, s_2$ which have similar wordings. Their dependency tree or 
semantic structure is extremely similar and that requires us to build a model 
which can recognize these subtle differences and reasoning to judge the 
sentence whether or not it makes sense.

\textbf{Subtask B:} Subtask B gives us one false sentence $s_f$(either $s_1$ or
$s_2$) which means this sentence does not make sense and three options $o_1, 
o_2, o_3$. We need to choose one right option which can explain why the give 
sentence does not make sense. 

\textbf{Subtask C:} Subtask C provides one false sentence $s_f$ as same as in 
subtask B and three references $r_1, r_2, r_3$. All these three references can 
explain why the false sentence does not make sense. This task requires us to 
build a model to generate the correct reason automatically given one false sentence.

\subsection{Pretrained Language Model}

\textbf{BERT} as a bidirectional pre-trained language model, 
has recently shown excellent performance in different downstream  tasks\cite{Devlin2019BERTPO}. It is an encoder based on multi-head attention with the self-attention mechanism in a fully connected layer. The input representation of BERT is constructed by summing the corresponding token, segment, and position embeddings. As an autoencoding (AE) model, It can capture the global context in both forward and backward directions. The pre-train of BERT uses two unsupervised strategies. 1) Masked LM; 2) Next Sentence Prediction (NSP). By optimizing for both of two tasks, BERT not only can learn semantic and synthetic knowledge but also world knowledge\cite{Rogers2020API}. These explain why BERT has astonishing performance.

\textbf{RoBERTa} is a extended study of BERT which showed that carefully 
tuning hyper-parameters and increase training data size lead to significantly improved results on language understanding. More specifically, \cite{Liu2019RoBERTaAR} proposed three methods to improve BERT 1) training the model longer, with bigger batches, over more data; 2) removing the next sentence prediction task; 3) training on longer sequences, and 4) dynamically changing the masking pattern applied to the training data. As same as other NLP tasks, RoBERTa gets more higher accuracy compared with BERT.

\section{Models}

Our proposed ERP system first generates (explain) its understanding of the given sentences by a language model, and then the explanation is used as a supplementary input to the prediction module. For subtask A the input is a sentence pair $s\_1$ and $s\_2$, and the input is the gainst-common-sense statement $s\_$ for subtask B. Subtask C is an explanation generation task and in this way, we could explore common-sense reasoning in two settings – 1) explain-and-then-predict and 2) predict-and-then-explain to evaluate the effectiveness of our ERP system. Therefore, we illustrate the ERP system consecutively for different sub-tasks in Sections 3.1 and 3.2. 

The architecture of our model is shown in Figure 1, the input x represents sequences (either one sentence or stacked sentences), and then for each token in this sequence is constructed by summing the corresponding token, segment and position embeddings. Then the semantic encoder map the input token into a vector in word-level (token-level), the transformer encoder captures the contextual information in sentence-level via the self-attention mechanism. After we get the contextual embedding vector, we use task-specific layer to apply downstream tasks, we use text classification layer here for both of task A and task B. We choose to introduce subtask A and subtask B first, and followed by subtask C for intuitive understanding, but for the competition, the organizer release the datasets of subtask A, subtask C, and subtask B in turn, which supports our ERP system.

\begin{figure}
    \centering
    \includegraphics[scale=0.3]{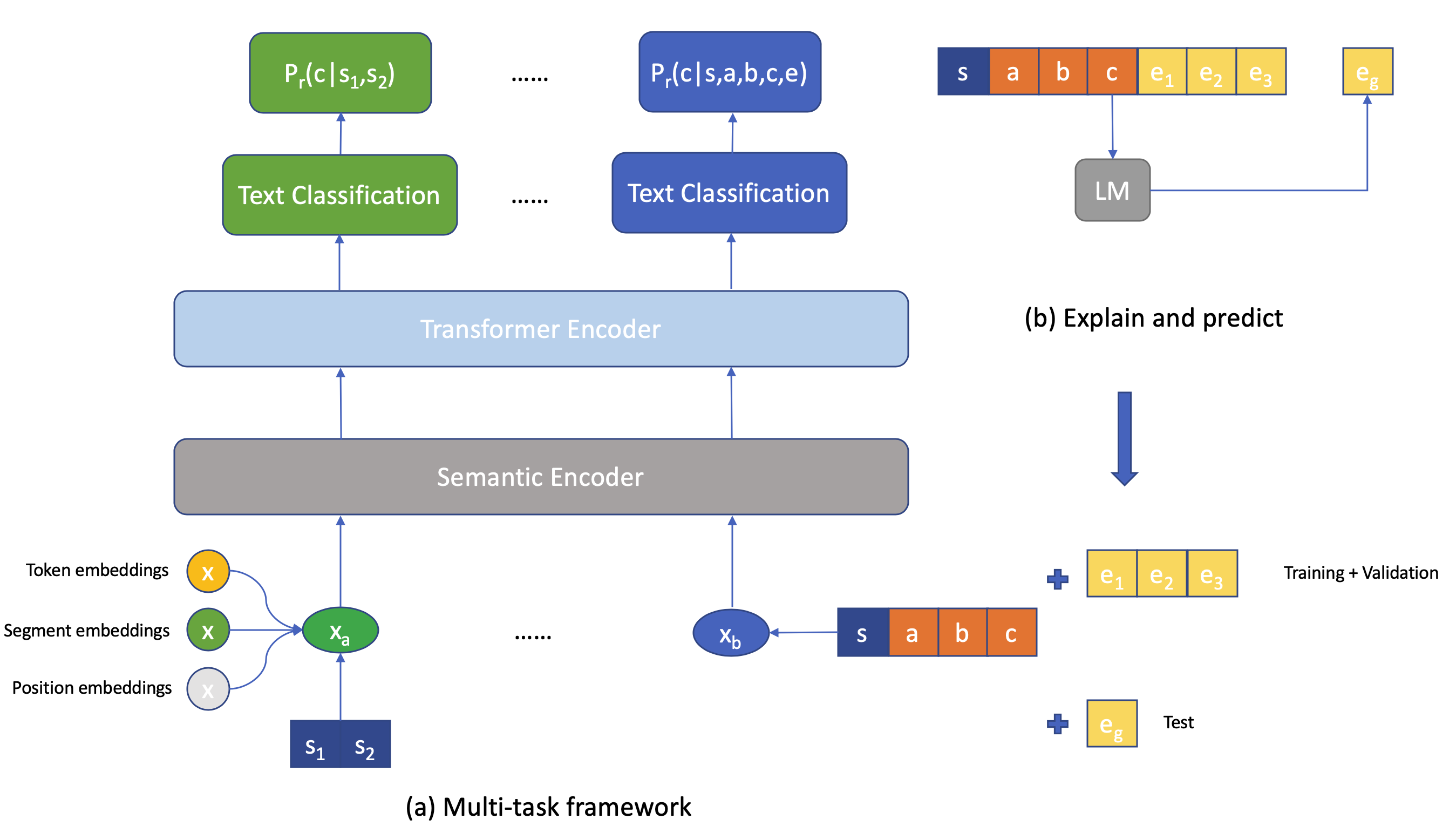}
    \caption{Explain, Reason and Predict (ERP) system, during traing and validation or test, subtask B uses different inputs.}
\end{figure}

\subsection{Sub-task A and Sub-task B}
For both of subtask A and subtask B, we cast them as text classification
problems. First of all, the training set of subtask A is $\Omega = \left\{ (s\_1^1, s\_2^1, y^1), (s\_1^2, s\_2^2, y^2), ..., (s\_1^N, s\_2^N, y^N) \right\} $, in which $s\_$ stands for two similar sentences and $y$ is label. In order to fine-tuning our model, we modified the input sequence x to $ x = ``[CLS] + s\_1 + [SEP] + s\_2" $, the [CLS] token is used for the final classfication, the [SEP] token is used to separate different sentences. 

Secondly, for subtask B, during training and validation, the organizer already release the correct explanation for each sent in subtask B, we need to use these data to generate the explanation for test data in subtask B, section 3.2 will introduce more details. Therefore, the generated explanation can be used to improve the performance of our model.
As shown in Figure 1, the input sequence consists of one false sent, three options, and some explanations (either ground-truth or generated) according to different periods. The training and validation sample can be cast as $\Omega = \left\{ (s^1, a^1, b^1, c^1, e_1^1, e_2^1, e_3^1), (s^2, a^2, b^2, c^2, e_1^2, e_2^2, e_3^2), ..., (s^N, a^N, b^N, c^N, e_1^N, e_2^N, e_3^N) \right\} $, the test samples are  $\Omega = \left\{ (s^1, a^1, b^1, c^1, e_g^1), (s^2, a^2, b^2, c^2, e_g^2), ..., (s^N, a^N, b^N, c^N, e_g^N) \right\} $. we still use the same structure to arrange our input but use additional special token to disambiguate different functions of sentences, like we use [OPTION] to represent three options, [EXP] to represent explanations. The objective of both subtask A and subtask B is to maximize: 

\begin{equation}
    L = \sum_{i=1}^{N}log p(y_i|x,\Theta) 
\end{equation}

\subsection{Sub-task C}
Here, we employ Commonsense Auto-Generated Explanations in \cite{wang-etal-2019-make}, generated by a language model. Subtask C provides one incorrect sentence and three references for explanation. All these three references can explain why the incorrect sentence does not make sense. Our LM is the large, pre-trained OpenAI GPT~\cite{radford2018improving}, which is a multi-layer transformer~\cite{vaswani2017attention} decoder. GPT is fine-tuned on the datasets. Thus, the input contains during the fine-tuning can be described as follows: 

\begin{equation}
    C_{ans} = ``s\ CUZ\ a, b, c"
\end{equation}

where the special token CUZ means ``is wrong may because". the input during testing is defined as follows:

\begin{equation}
    C_{ans} = ``s, CUZ\ ?"
\end{equation}

The model is trained to generate the explanation $e$ on the basis of conditional language modeling objective, the goal is to maximize:

\begin{equation}
    \sum_{i}{log P (e_i|e_{i-k},...,e_{i-1},C_{ans};\theta)}
\end{equation}
where k is the size of the context window. The conditional probability $P$ is modeled by a neural network with parameters $\Theta$ conditioned on $C_{ans}$ and previous explanation tokens.

\section{Experiment}
It is important to make it clear that all our experiments are conducted which meet the requirement of the competition. We can not use the dataset which is not released during the formal competition which means we can not use subtask B data for subtask A and subtask C, because subtask B is released at last, etc.

\subsection{Baseline of Sub-task A and Subtask B}
As described before, the project consists of three subtasks. Subtask A 
is to choose from two natural language statements with similar wordings 
which one makes sense and another one does not make sense. Subtask 
B is to find the key reason why a given statement does not make sense. 
Subtask C asks the machine to generate reasons. Subtask A and B are 
evaluated by accuracy and Subtask C is evaluated using BLEU. The organizers use a random subset of the test set and will do a human evaluation to further evaluate the 
systems with a relatively high BLEU score (which is not conducted in the post-evaluation period).

First of all, we use BERT and RoBERTa as our baseline since both of them show impressive performance in many NLP downstream tasks. Table 1 shows results compare BERT with RoBERTa that use different corpus for each task. For subtask A, the RoBERTa model reaches the highest accuracy 86.2\% in the test and 88.5\% in dev datasets. For subtask B, when we add data from subtask A, the performance get the peak at 82.3\% accuracy, but it attracts our attention when we use additional subtask C data that the dev accuracy is extremely high with the test accuracy is obviously lower. We assume 1) the data from subtask C show tremendous potential ability to solve subtask B 2) the model relies too much on Subtask C data, resulting in very low performance without it. After we use the generated explanation during the test, the model gets considerable improvement which validates our assumption. 

\begin{table}[ht]
    \centering
    \begin{tabular}{c|l|c|c|c}
    \toprule[1pt]
    Model & Task & Test & Dev & Label  \\
    \hline
    \multirow{4}*{BERT} & Task A & 85.3\% & 85.9\% & 2 \\
        \cline{2-5}
        & Task B & 79.1\% & 79.7\% & 3 \\
        & \quad + Task A & 73.4\% & 82.4\% & 3 \\
        & \quad + Task C & $51.5\%^1, 54.6\%^2$ & 82.5\% & 3 \\
    \hline
    \multirow{4}*{RoBERTa} & Task A & \textbf{86.2\%} & \textbf{88.5\%} & 2 \\
        \cline{2-5}
        & Task B & 81.4\% & 84.6\% & 3 \\
        & \quad + Task A & \textbf{82.3\%} & 84.5\% & 3 \\
        & \quad + Task C & $46.5\%^1, 48.9\%^2$ & \textbf{99.9\%} & 3 \\
    \bottomrule[1pt]
    \end{tabular}
    \caption{Task A and B: Baseline Results, $51.5\%^1$ represents we do
    not have explanations during test, but $54.6\%^2$ means we use our 
    generated explanations, etc.}
\end{table}

\subsection{Explain and Predict}
To better understand this deviant phenomenon, we present results with different sample percentages when we randomly choose whether or not to use the subtask C data which is shown at Table 2. Specifically, under the condition of 7:3 sample percent, when we get a sample from subtask B, and then we choose to inject additional explanation with a 30\% probability, but not with 70\% probability. If we decide to inject additional knowledge, then we will sample one, two, and three explanations with the equal possibility. We observe that the accuracy of dev datasets becomes a little lower, but the test accuracy gets comparable improvement. We force the model to learn more with limited external knowledge through this approach, and the result further validates our assumption before. This leads to the appearance of our ``Explain, Reason and Predict” (ERP) system and provides an interpretable foundation. Then all we need to do is to improve our baseline, in which we try different ways such as knowledge inject and multi-tasks.

\begin{table}[ht]
    \centering
    \begin{tabular}{c|l|c|c}
    \toprule[1pt]
    Sample Percent & Corpus & Test & Dev \\
    \hline
    \multirow{3}*{5:5} & Single Exp & \textbf{80.6\%} & \textbf{89.9\%} \\
        \cline{2-4}
        & \qquad + Task A & 80.5\% & 86.6\% \\
        & All Exp + Task A & 80.0\% & 88.2\% \\
    \hline
    \multirow{3}*{7:3} & Single Exp & \textbf{82.3\%} & \textbf{87.1\%} \\
        \cline{2-4}
        & \qquad + Task A & 81.1\% & 85.5\% \\
        & All Exp + Task A & 81.9\% & 86.6\% \\
    \bottomrule[1pt]
    \end{tabular}
    \caption{Rational Experiment on Task B, 5:5 means we use 50\% of samples to train with exp, 50\% to train without exp, etc.}
\end{table}

\subsection{Multi-task}
During the experiment, we find an interesting case illustrating that multi-task learning may help a lot at sub-tasks A and B. Given a false sentence s: “an umbrella can help you keep warm in snowy days.” and three options: A. “we don't wear umbrellas”, B: “umbrellas can keep you dry in snowy days”, C: “going outside is very crazy in snowy days”. The ground truth is A, but the model outputs B which can better explain the sentence. After we check the other true sentence “a thicker cloth can help you keep warm in snowy days” in subtask A dataset, we know why the ground truth is A. Obviously, we need some knowledge at subtask A to help us to solve task B, so we think multi-tasks learning is a direction worthy to try~\cite{Liu2019MultiTaskDN}.

Rather than enriching semantic embedding with knowledge graph, we leverage existing
datasets across different domains which also require common sense reasoning like 
ARC, CommonseQA, and so on. We believe multi-task learning can learn more robust and
universal embedding and then make our model get better performance and improve our 
baseline. In the following experiments, we will validate including additional 
datasets as external input information can boost our performance of our ERP system.

\begin{table}[ht]
    \centering
    \begin{tabular}{c|l|c|c}
    \toprule[1pt]
    Task & Model & Test & Dev \\
    \hline
    \multirow{3}*{Task A} & MT-SAN on Task A (single task) & 91.7\% & 94.8\% \\
        \cline{2-4}
        & MT-SAN on Task A+MNLI+SciTail+MRPC & 92.8\% & 94.8\% \\
        \cline{2-4}
        & MT-SAN (ensemble) & \textbf{92.9\%} & \textbf{95.1\%} \\ 
    \hline
    \multirow{3}*{Task B} & MT-SAN on Task B (single task) & 87.3\% & 88.1\% \\
        \cline{2-4}
        & MT-SAN on Task B+ARC+CommonseQA & $89.6\%, 89.3\%^*$ & 91.0\%, $\textbf{93.5\%}^{*}$ \\
        \cline{2-4}
        & MT-SAN (ensemble) & \textbf{89.7\%} & 92.2\% \\ 
    \bottomrule
    \end{tabular}
    \caption{Multi-Task Result, $93.5\%^*$ means Explain and Predict we said before.}
\end{table}

Table 3 shows the results obtained by our final Multi-task ERP model. We report our two best models that ensemble models using different dropout rates, see more details in the following section. At subtask A, our ensemble model reaches 92.9\% accuracy and 95.1\% accuracy during test and dev respectively, while getting 89.7\% accuracy at the test of subtask B and 93.5\% accuracy at dev of subtask B. The highest accuracy in dev dataset of subtask B indicates the tremendous potential of our ERP system. Using additional datasets together to train provides marginal improvement compare with a single task which attributes to better model generalization under multi-task setting from our point of view.

\subsection{Implement details}
Our implementation of Multi-Task ERP system is based on \cite{Liu2019MultiTaskDN}. We used Adamax as our optimizer with a learning rate of 5e-5 and a batch size of 4.We set the number of epochs of 10 and use a linear learning rate decay schedule with warm-up over 0.1. We also set the dropout rate of all the task-specific layers as 0.1, except for ensemble models which we set different dropout rates to get different models. According to \cite{Liu2019ImprovingMD}, we set dropout rate ranged in $\left\{0.1,0.2,0.3 \right\}$. To avoid the exploding gradient problem, we clipped the gradient norm within 1. We set the mixture ratio as 0.4 to re-weighting different tasks\cite{Xu2018MultitaskLW}, more details can be found in the implementation \footnote{The codes are available at: https://github.com/ruleGreen/myMTDNN}.

\subsection{Subtask C}
Since this is a text generation problem, we choose to use the GPT model as our baseline. Since some of the samples use knowledge from subtask A, we conducted contrast experiments by using data from subtask A and CoS-E\cite{Rajani2019ExplainYL}. We observed that adding explanations led to a very small decrease in the performance compared to the baseline at test datasets, but adding data from subtask A improve about 0.3. Some generated explanations are showed in Table 5 below.

\begin{table}[ht]
    \centering
    \begin{tabular}{c|l|c|c}
    \toprule
    Model & Corpus & Test & Dev \\
    \hline
    \multirow{3}*{GPT} & Task C & 12.65 & 5.96 \\
         & \quad + Task A & \textbf{12.94} & 5.99 \\
         & \quad + Aug  & 12.31 & \textbf{6.54} \\
    \bottomrule
    \end{tabular}
    \caption{Task C: CommonSense Explanation}
\end{table}

\begin{table}[ht]
    \centering
    \begin{tabular}{l|l}
    \toprule
    \begin{tabular}[c]{@{}l@{}}False Sent\\ True Sent\end{tabular} & Generated Exp \\
    \hline
    \begin{tabular}[c]{@{}l@{}}The inverter was able to power the house\\ The inverter was able to power the house\end{tabular} & The inverter was able to power the house \\
    \hline
    \begin{tabular}[c]{@{}l@{}}There are beautiful planes here and there in the garden\\ There are beautiful flowers here and there in the garden\end{tabular} & planes are not found in the garden. \\
    \hline
    \begin{tabular}[c]{@{}l@{}}The chef put extra lemons on the pizza.\\ The chef put extra mushrooms on the pizza.\end{tabular} & the chef does not put extra lemons on the pizza. \\
    \hline
    \begin{tabular}[c]{@{}l@{}}She is hypnotizing a book.\\ She is reading a book.\end{tabular} & books are not hypnotizing. \\
    \hline
    \begin{tabular}[c]{@{}l@{}}The boy ate a basketball\\ The boy ate a donut\end{tabular} & basketballs are not edible. \\
    \bottomrule
    \end{tabular}
    \caption{Task C: Generated Exp(Also used in explain and predict part)}
\end{table}




Compared with the original paper\cite{wang-etal-2019-make}, our model gets much higher accuracy in both of subtask A and subtask B. Our performance rank 10th on 29 April 2019, with 92.9\% accuracy at subtask A (rank 11), 89.7\% at subtask B (rank 9), 12.9 at subtask C (rank 8)\footnote{All results before 29 April, 2020}.

\section{Analysis}
Despite the strong performance of our model, it still fails to detect some samples at subtask A and subtask B, and few sentences generated by our model can not well explain why the given sentence does not make sense. An in-depth analysis of these samples shows that they can be clustered into some classes.

\subsection{Error Analysis at Subtask A}

\begin{itemize}
	\item \textbf{Basic common sense knowledge} which can be solved by introduce external knowledge graph like ConceptNet (eg., $s\_1$: The moon sets at night, $s\_2$: The sun sets at night, label: 1, prediction: 2).
	\item \textbf{Implicit common sense knowledge.} Current knowledge graphs do not contain everything about common sense knowledge because the limitation of memory and the huge volume of common sense, and it still needs better solutions by using more comprehensive knowledge representation and transfer learning or other methods (eg., $s\_1$: Cats have got seven lives, $s\_2$: Cats have got one life, label: 1, prediction: 2).
	\item \textbf{Specific domain knowledge} required to make a correct judgment (eg., $s\_1$: Hair is already dead, $s\_2$: Hair screams when you cut it, label: 2, prediction: 1), since the human may not know that the hair is dead protein cells, so it is a big challenge for the model to learn this rare domain knowledge from large corpus and datasets.
	\item \textbf{Others} (eg., $s\_1$: coffee takes sleep, $s\_2$: coffee depresses people, label: 2, prediction: 1; $s\_1$: The sun is black, $s\_2$: The sun is white, label: 1, prediction: 2)
\end{itemize}

\subsection{Error Analysis at Subtask B}


For subtaskB, there are two different ways to address it: the conventional and the explain and predict methods, that leads three different cases 1) both methods are wrong 2) the conventional one is correct but the other wrong 3) ERP system is correct but the other wrong. According to a comprehensive analysis below, we find that our ERP system can reason and make a more persuasive decision than the conventional one. 

\begin{enumerate}
	\item  Both methods output the wrong judgment
    \begin{enumerate}
    	\item  \textbf{Explain in different perspectives or levels} (eg., $s_f$: Everyone loves reading horror novels. $o_1$: Horror novels are scary. $o_2$: Reading novels can be a good way to relax. $o_3$: Not everyone likes to read horror
novels. label: C, prediction: A). Why the given $s_f$ does not make sense can have 
multiple explanations in different levels. Here, to explain why ``Everyone loves 
reading horror novels" does not make sense, from our point of view, both $o_1$ and 
$o_3$ are correct if we assume the given $s_f$ is already false, since they can 
composite ``$s_f$ is wrong because $o_1$ or $o_3$". We think $o_1$ gives 
explanation from more subtle and deeper level than $o_3$.
        \item \textbf{Implicit 
common sense knowledge} as same as in subtask A (eg., $s_f$: drama plays are often 
performed before cows, $o_1$: this rural drama tells the story of a cow, $o_2$: the
cow is a kind of animal while drama isn't, $o_3$: a cow is unable to appreciate and
understand the drama, label: C, prediction: B), we need to know that drama plays 
are appreciated and understand by people in this example.
    \end{enumerate}
    \item Examples classified wrongly by the conventional methods but not our ERP system
    \begin{enumerate}
    	\item \textbf{Lack of reasoning capability} which equipped in our ERP system (eg., $s_f$: shoes can fly, $o_1$: There are many creatures that can fly, $o_2$: Shoes do not have wings, $o_3$: People cannot fly, label: B, prediction: C). The conventional can not reason those wings are needed to fly here.
        \item \textbf{Basic common sense knowledge}. The model still needs external knowledge to support making the right classification.
    \end{enumerate}
    \item About 1.10\% of samples are not classified correctly by our model but the conventional ones, we think this mostly attribute to noise introduced by multi-task setting.
    \begin{enumerate}
    	\item \textbf{Capture plausible knowledge} (eg., $s_f$: the lava was warm and soft, $o_1$: lava can destroy the warm and soft cake, $o_2$: 
lava is too hard to be soft, $o_3$: lava is too hot to be warm or soft, label: C, prediction: B). The model captured that something is too hard to be soft, but it ignores the attributes of lava. 
        \item \textbf{Others} (eg., $s_f$: it is said that Santa comes on Thanksgiving Days, $o_1$: Santa comes on Christmas day, not Thanksgiving Day, $o_2$: Santa is a figure in legend, not reality, $o_3$: Santa is a figure in western culture, not eastern culture, label: A, prediction: B). We first think this is caused by explaining in different perspectives or levels which described above, but after we check the whole data, and we find an example ($s_f$: Santa Claus sent Jim a Christmas present, $o_1$: There aren't Santa Claus in the world, $o_2$: Santa Claus is very busy, $o_3$: Santa Claus is old, label: A) in the training data of subtask B. This proves that our model can learn more robust and universal embedding than the conventional method.
    \end{enumerate}
\end{enumerate}




\subsection{Error Analysis at Subtask C}

Although most of the results make sense, but there are still some generated reasons which can not well explain why the given sentence does not make sense. Most cases, as we found, are with: 
\begin{itemize}
	\item \textbf{Wrong explain direction} (eg., $s_f$: The 
inverter was able to power the continent, $e_g$: inverter is not a living thing). 
    \item \textbf{Repetition} (eg., $s_f$: sugar is used to make coffee sour, $e_g$: sugar is used to make coffee). Like the example, some cases contain repeatedly generated words.
\end{itemize}

\section{Conclusion}
\label{chap:chapter-6}
In this paper we present our model on the task
of Commonsense Validation and Explanation (ComVE) in
SemEval-2020. We explore multi-task learning to jointly learn how to inference the hidden common-sense fact and do common-sense reasoning with the RoBERTa model and
achieved competitive results. 
Our result analysis indicates that our ”Explain, Reason and Predict” approach helps improve the performance of RoBERTa and have a strong reasoning capability. The biggest regret in this competition is that we did not incorporated with world knowledge by introducing some knowledge graphs. Due to implicit common sense knowledge and different explanation perspectives, we still need more efforts on the model architecture and find more elegant ways to inject knowledge. Our positive results point to future work in extending the ERP approach to a variety of other types of
common sense reasoning tasks.


\section*{Acknowledgement}
This work is partly supported by Hong Kong RGC GRF (14204118) and ITF (ITS/335/18). We thank the three anonymous reviewers for the insightful suggestions on various
aspects of this work.



\bibliographystyle{coling}
\bibliography{semeval2020}

\begin{thebibliography}{}

\bibitem[\protect\citename{Asher and Vieu}1995]{asher1995toward}
Nicholas Asher and Laure Vieu.
\newblock 1995.
\newblock Toward a geometry of common sense: A semantics and a complete
  axiomatization of mereotopology.
\newblock In {\em IJCAI (1)}, pages 846--852.

\bibitem[\protect\citename{Bosselut \bgroup et al.\egroup
  }2019]{Bosselut2019COMETCT}
Antoine Bosselut, Hannah Rashkin, Maarten Sap, Chaitanya Malaviya, Asli
  Çelikyilmaz, and Yejin Choi.
\newblock 2019.
\newblock Comet: Commonsense transformers for automatic knowledge graph
  construction.
\newblock In {\em ACL}.

\bibitem[\protect\citename{Devlin \bgroup et al.\egroup
  }2019]{Devlin2019BERTPO}
Jacob Devlin, Ming-Wei Chang, Kenton Lee, and Kristina Toutanova.
\newblock 2019.
\newblock Bert: Pre-training of deep bidirectional transformers for language
  understanding.
\newblock {\em ArXiv}, abs/1810.04805.

\bibitem[\protect\citename{Forbes and Choi}2017]{forbes2017verb}
Maxwell Forbes and Yejin Choi.
\newblock 2017.
\newblock Verb physics: Relative physical knowledge of actions and objects.
\newblock In {\em ACL}.

\bibitem[\protect\citename{He \bgroup et al.\egroup }2019]{He2019AHN}
Pengcheng He, Xiaodong Liu, Weizhu Chen, and Jianfeng Gao.
\newblock 2019.
\newblock A hybrid neural network model for commonsense reasoning.
\newblock {\em ArXiv}, abs/1907.11983.

\bibitem[\protect\citename{Lin \bgroup et al.\egroup }2019]{Lin2019KagNetKG}
Bill~Yuchen Lin, Xinyue Chen, Jamin Chen, and Xiang Ren.
\newblock 2019.
\newblock Kagnet: Knowledge-aware graph networks for commonsense reasoning.
\newblock In {\em EMNLP/IJCNLP}.

\bibitem[\protect\citename{Liu \bgroup et al.\egroup
  }2019a]{Liu2019ImprovingMD}
Xiaodong Liu, Pengcheng He, Weizhu Chen, and Jianfeng Gao.
\newblock 2019a.
\newblock Improving multi-task deep neural networks via knowledge distillation
  for natural language understanding.
\newblock {\em ArXiv}, abs/1904.09482.

\bibitem[\protect\citename{Liu \bgroup et al.\egroup
  }2019b]{Liu2019MultiTaskDN}
Xiaodong Liu, Pengcheng He, Weizhu Chen, and Jianfeng Gao.
\newblock 2019b.
\newblock Multi-task deep neural networks for natural language understanding.
\newblock In {\em ACL}.

\bibitem[\protect\citename{Liu \bgroup et al.\egroup }2019c]{Liu2019RoBERTaAR}
Yinhan Liu, Myle Ott, Naman Goyal, Jingfei Du, Mandar Joshi, Danqi Chen, Omer
  Levy, Mike Lewis, Luke Zettlemoyer, and Veselin Stoyanov.
\newblock 2019c.
\newblock Roberta: A robustly optimized bert pretraining approach.
\newblock {\em ArXiv}, abs/1907.11692.

\bibitem[\protect\citename{Peters \bgroup et al.\egroup }2018]{Peters:2018}
Matthew~E. Peters, Mark Neumann, Mohit Iyyer, Matt Gardner, Christopher Clark,
  Kenton Lee, and Luke Zettlemoyer.
\newblock 2018.
\newblock Deep contextualized word representations.
\newblock In {\em Proc. of NAACL}.

\bibitem[\protect\citename{Radford \bgroup et al.\egroup
  }2018]{radford2018improving}
Alec Radford, Karthik Narasimhan, Tim Salimans, and Ilya Sutskever.
\newblock 2018.
\newblock Improving language understanding by generative pre-training.
\newblock {\em URL https://s3-us-west-2. amazonaws.
  com/openai-assets/researchcovers/languageunsupervised/language understanding
  paper. pdf}.

\bibitem[\protect\citename{Rajani \bgroup et al.\egroup
  }2019]{Rajani2019ExplainYL}
Nazneen~Fatema Rajani, Bryan McCann, Caiming Xiong, and Richard Socher.
\newblock 2019.
\newblock Explain yourself! leveraging language models for commonsense
  reasoning.
\newblock In {\em ACL}.

\bibitem[\protect\citename{Rogers \bgroup et al.\egroup }2020]{Rogers2020API}
Anna Rogers, Olga Kovaleva, and Anna Rumshisky.
\newblock 2020.
\newblock A primer in bertology: What we know about how bert works.
\newblock {\em ArXiv}, abs/2002.12327.

\bibitem[\protect\citename{Speer \bgroup et al.\egroup
  }2016]{Speer2016ConceptNet5A}
Robyn Speer, Joshua Chin, and Catherine Havasi.
\newblock 2016.
\newblock Conceptnet 5.5: An open multilingual graph of general knowledge.
\newblock In {\em AAAI}.

\bibitem[\protect\citename{Talmor \bgroup et al.\egroup
  }2019]{Talmor2019CommonsenseQAAQ}
Alon Talmor, Jonathan Herzig, Nicholas Lourie, and Jonathan Berant.
\newblock 2019.
\newblock Commonsenseqa: A question answering challenge targeting commonsense
  knowledge.
\newblock {\em ArXiv}, abs/1811.00937.

\bibitem[\protect\citename{Vaswani \bgroup et al.\egroup
  }2017]{vaswani2017attention}
Ashish Vaswani, Noam Shazeer, Niki Parmar, Jakob Uszkoreit, Llion Jones,
  Aidan~N Gomez, {\L}ukasz Kaiser, and Illia Polosukhin.
\newblock 2017.
\newblock Attention is all you need.
\newblock In {\em Advances in neural information processing systems}, pages
  5998--6008.

\bibitem[\protect\citename{Wang \bgroup et al.\egroup
  }2019a]{wang-etal-2019-make}
Cunxiang Wang, Shuailong Liang, Yue Zhang, Xiaonan Li, and Tian Gao.
\newblock 2019a.
\newblock Does it make sense? and why? a pilot study for sense making and
  explanation.
\newblock In {\em Proceedings of the 57th Annual Meeting of the Association for
  Computational Linguistics}, pages 4020--4026, Florence, Italy, July.
  Association for Computational Linguistics.

\bibitem[\protect\citename{Wang \bgroup et al.\egroup
  }2019b]{Wang2019ImprovingNL}
Xiaoyan Wang, Pavan Kapanipathi, Ryan Musa, Mo~Yu, Kartik Talamadupula, Ibrahim
  Abdelaziz, Maria Chang, Achille Fokoue, Bassem Makni, Nicholas Mattei, and
  Michael~J. Witbrock.
\newblock 2019b.
\newblock Improving natural language inference using external knowledge in the
  science questions domain.
\newblock In {\em AAAI}.

\bibitem[\protect\citename{Wang \bgroup et al.\egroup
  }2020]{wang-etal-2020-semeval}
Cunxiang Wang, Shuailong Liang, Yili Jin, Yilong Wang, Xiaodan Zhu, and Yue
  Zhang.
\newblock 2020.
\newblock {S}em{E}val-2020 task 4: Commonsense validation and explanation.
\newblock In {\em Proceedings of The 14th International Workshop on Semantic
  Evaluation}. Association for Computational Linguistics.

\bibitem[\protect\citename{Xu \bgroup et al.\egroup }2018]{Xu2018MultitaskLW}
Yichong Xu, Xiaodong Liu, Yelong Shen, Jingjing Liu, and Jianfeng Gao.
\newblock 2018.
\newblock Multi-task learning with sample re-weighting for machine reading
  comprehension.
\newblock In {\em NAACL-HLT}.

\bibitem[\protect\citename{Zhong \bgroup et al.\egroup
  }2019]{Zhong2019KnowledgeEnrichedTF}
Peixiang Zhong, Di. Wang, and Chunyan Miao.
\newblock 2019.
\newblock Knowledge-enriched transformer for emotion detection in textual
  conversations.
\newblock In {\em EMNLP/IJCNLP}.

\bibitem[\protect\citename{Zhou \bgroup et al.\egroup
  }2019]{zhou2019evaluating}
Xuhui Zhou, Yue Zhang, Leyang Cui, and Dandan Huang.
\newblock 2019.
\newblock Evaluating commonsense in pre-trained language models.
\newblock {\em arXiv preprint arXiv:1911.11931}.

\end{thebibliography}

\end{document}